\documentclass{article}

\usepackage{arxiv}
\usepackage{amsmath,amssymb,amsfonts}
\usepackage{algorithmic}
\usepackage{graphicx}
\usepackage{multirow}
\usepackage{adjustbox}
\usepackage{array}
\usepackage[section]{placeins}
\usepackage{xcolor}
\definecolor{darkgreen}{rgb}{0.0, 0.5, 0.0}
\usepackage{algorithm}
\usepackage{algorithmic}
\usepackage{stfloats}
\usepackage{booktabs}
\usepackage{textcomp}
\usepackage{multicol}
\usepackage{array}
\usepackage{threeparttable}

\usepackage[accsupp]{axessibility}  

\usepackage{float}

\usepackage[colorlinks=true, allcolors=blue]{hyperref}

\title{RobMOT: Robust 3D Multi-Object Tracking by Observational Noise and State Estimation Drift Mitigation on LiDAR PointCloud}

\author{
 Mohamed Nagy \\
  Electrical and Computer Engineering\\
  Khalifa University\\
  \texttt{mohamed.nagy@ieee.org} \\
   \And
 Naoufel Werghi \\
  Electrical and Computer Engineering\\
  Khalifa University\\
  \texttt{naoufel.werghi@ku.ac.ae} \\
  \And
 Bilal Hassan \\
  Electrical and Computer Engineering\\
  Khalifa University\\
  \texttt{bilal.hassan@ku.ac.ae} \\
  \And
   Jorge Dias \\
  Electrical and Computer Engineering\\
  Khalifa University\\
  \texttt{jorge.dias@ku.ac.ae} 
   \And
   Majid Khonji \\
  Electrical and Computer Engineering\\
  Khalifa University\\
  \texttt{majid.khonji@ku.ac.ae} 
}

\begin{document}
\maketitle
\begin{abstract}
This paper addresses key limitations in recent 3D tracking-by-detection methods,  focusing on the challenges of identifying legitimate trajectories and mitigating state estimation drift in the Kalman filter. Current methods rely heavily on threshold-based filtering of detection scores approach to reduce false positives and prevent ghost trajectories. However, this approach fails for distant and partially occluded objects, where detection scores drop, and false positives surpass that threshold. Additionally, many existing methods assume detections offer precise localization, overlooking inherent noise, which affects localization accuracy and causes state drift for occluded objects, as demonstrated in this work. To this end, a novel track validity mechanism, combined with a multi-stage observational gating process, is proposed that significantly reduces ghost tracks and improves tracking performance. Our method achieves $29.47\%$ enhancement in Multi-Object tracking accuracy (MOTA) on the KITTI validation dataset with the Second detector. Furthermore, a refined Kalman filter term mitigates localization noise, ensuring robust state estimation for occluded objects and superior recovery during prolonged occlusions, with higher order tracking accuracy (HOTA) improving by $4.8\%$ on the KITTI validation dataset with PV-RCNN detector. The proposed online framework, RobMOT, outperforms state-of-the-art methods, including deep learning approaches, across multiple detectors, with HOTA improvements of up to $3.92\%$ on the KITTI testing dataset and $8.7\%$ on the KITTI validation dataset while achieving the lowest identity switch (IDSW) scores of $7$ and $0$, respectively. RobMOT excels under challenging scenarios, such as tracking distant objects and handling prolonged occlusions, surpassing state-of-the-art methods on the Waymo Open testing dataset with a $1.77\%$ improvement in MOTA for objects at distances exceeding $50$ meters. RobMOT achieves a groundbreaking run-time of 3221 FPS using a single CPU, establishing itself as a highly efficient and scalable solution for real-time multi-object tracking.
  
\end{abstract}
\section{Introduction}
\label{sec:intro}
\begin{figure}[t]
  \centering
  \includegraphics[width=\linewidth]{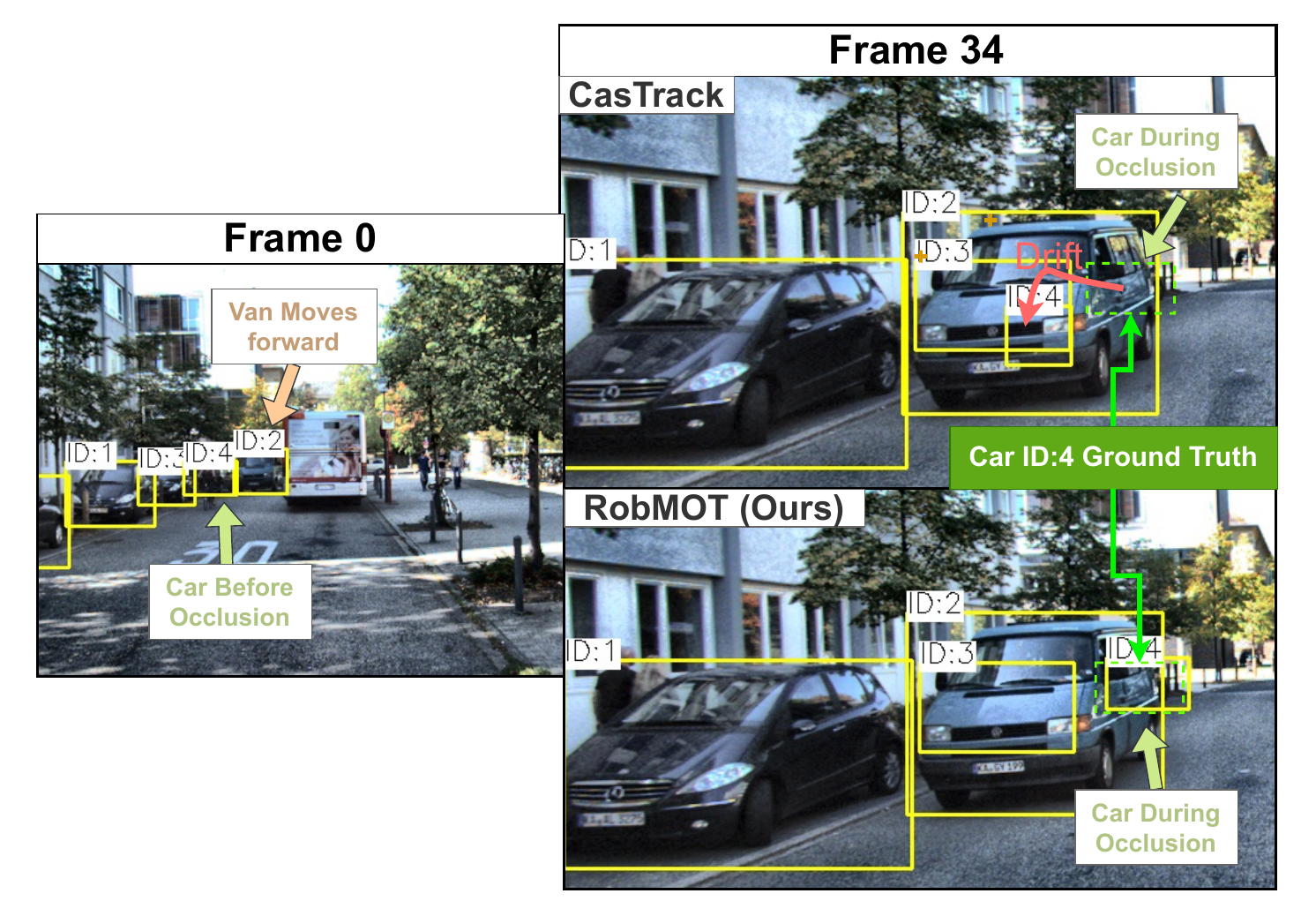}
  \caption{A trajectory drift scenario involving a car (ID4)  hidden by a van (ID2). (a) At frame 0, four cars are detected and identified (ID1, ID2, ID3, ID4). (b) By frame 34, the state-of-the-art method~\cite{CasTrack, VirConvDet_CVPR_23} incorrectly estimates the position of car ID4 due to state estimation drift. A dashed green rectangle indicates the actual position. (c) Our proposed approach, with its noise mitigation techniques, successfully localizes car ID4 accurately.}
  \label{fig:trajectory_drift_scene}
  \vspace{-0.6cm}
\end{figure}
3D multi-object tracking (MOT) is one of the pillars of autonomous driving. It is responsible for determining the state of objects in the environment, including their spatial location and motion parameters like velocity and acceleration. Wu et al.~\cite{Wu2023Supplemental, CasTrack}, propose an online MOT method, CasTrack, with CasA detector~\cite{casA_detector} that outperforms the recently published benchmarks, according to the KITTI leaderboard~\cite{Geiger2012CVPR}. In another experiment, the author employs the same tracking method with the VirConv detector~\cite{VirConvDet_CVPR_23} named VirConvTrack~\cite{Wu2023Supplemental}.  However, trajectory drift has been observed in state estimation of occluded objects in their work, as shown in Figure~\ref{fig:trajectory_drift_scene}, preventing MOT methods from recovering objects under challenging tracking conditions like prolonged occlusion and distant objects as emphasized in Figure~\ref{fig:occlusion_comp}. Our investigation shows that the drift is caused by cumulative inaccuracies in bounding box localization from deep learning models. These models produce detection bounding boxes that slightly shift around an object's true position in LiDAR data over time. This spatial localization noise creates an illusion of motion, ultimately affecting the object's state estimation accuracy.\\
Another issue is the absence of trajectory legitimacy verification in the MOT literature~\cite{CasTrack, deepfusion_mot, intro_tra_by_det_kim_eager, intro_false_positive, Rethink_mot, oc_sort}. These works apply a fine-tuned threshold on detection scores of the detections provided by deep learning models to filter out false positive observations that cause ghost trajectories. This approach is not reliable as it can block observations that belong to actual (legitimate) trajectories while leaking false positive observations. Since observations of legitimate trajectories can inherit some of the false positive characteristics, such as low detection scores for distant and partial occlusion scenarios, the MOT methods need to validate trajectories based on the consistency and legitimacy of their historical observations instead.
\\ This work tackles the above-mentioned challenges. First, a refined Kalman filter (KF) is introduced that mitigates the spatial localization noise responsible for trajectory drift in Section~\ref{subsec:trajectory_ass_state_estm}. Second, a novel online mechanism is introduced that validates and distinguishes between legitimate and ghost trajectories based on studying the characteristics of ghost tracks, discussed in Section~\ref{subsec:Trajectory_Validity}. Our proposed refinement shows consistency in tracking for occluded objects superior to the latest state-of-the-art approach~\cite{CasTrack, Wu2023Supplemental} demonstrated in Figure~\ref{fig:trajectory_drift_scene} and Figure~\ref{fig:occlusion_comp}. Next, the proposed trajectory validity mechanism incorporates a multi-stage observational gate that allows potential detections of legitimate tracks to pass to the system and an uncertainty formulation used to quantify the temporal improvement of trajectory certainty based on ghost track characteristics. On the KITTI validation dataset~\cite{Geiger2012CVPR}, the mechanism shows a $8.93\%$ and $29.47\%$ enhancement in higher order tracking accuracy (HOTA) and Multi-Object tracking accuracy (MOTA) metrics with Second detector~\cite{second_detector}, respectively. \\
RobMOT is an online 3D MOT framework that integrates the above-mentioned innovations.
Our contribution can be summarized as follows:
\begin{enumerate}
    \item To the best of our knowledge, this is the first work to uncover the trajectory drift noise associated with detections and its impact on the state estimation of occluded objects in 3D MOT. \textit{Our refinement in KF shows superior performance over the top-performing benchmark~\cite{CasTrack, Wu2023Supplemental} on challenging  MOT conditions, demonstrated in Figure~\ref{fig:occlusion_comp}}.
    \item A novel online trajectory validity mechanism is proposed for temporal distinguishing between ghost and legitimate tracks, supported by a multi-stage observational gate for potential observation identification of the legitimate tracks. \textit{The mechanism reduces ghost tracks that improve HOTA by $8.93\%$ and MOTA by $29.47\%$ with Second detector~\cite{second_detector}, as demonstrated in the ablation study Section~\ref{subusbsec:track_val_and_gho_trak}.}
    \item Lastly, Robmot is an online 3D MOT framework with exceptional tracking performance, surpassing recent benchmarks to show its robustness across different datasets and detectors. \textit{The tracking performance margin of our framework with the state-of-the-art~\cite{CasTrack, Wu2023Supplemental} reaches $8.7\%$ in HOTA and $16.1\%$ in MOTA on KITTI validation dataset, and significant enhancement in computational latency as demonstrated in Figure~\ref{fig:abl_speed}.}
\end{enumerate}

\section{Related Work}
\begin{figure*}[tb]
  \centering
  \includegraphics[width=\linewidth,height=9cm]{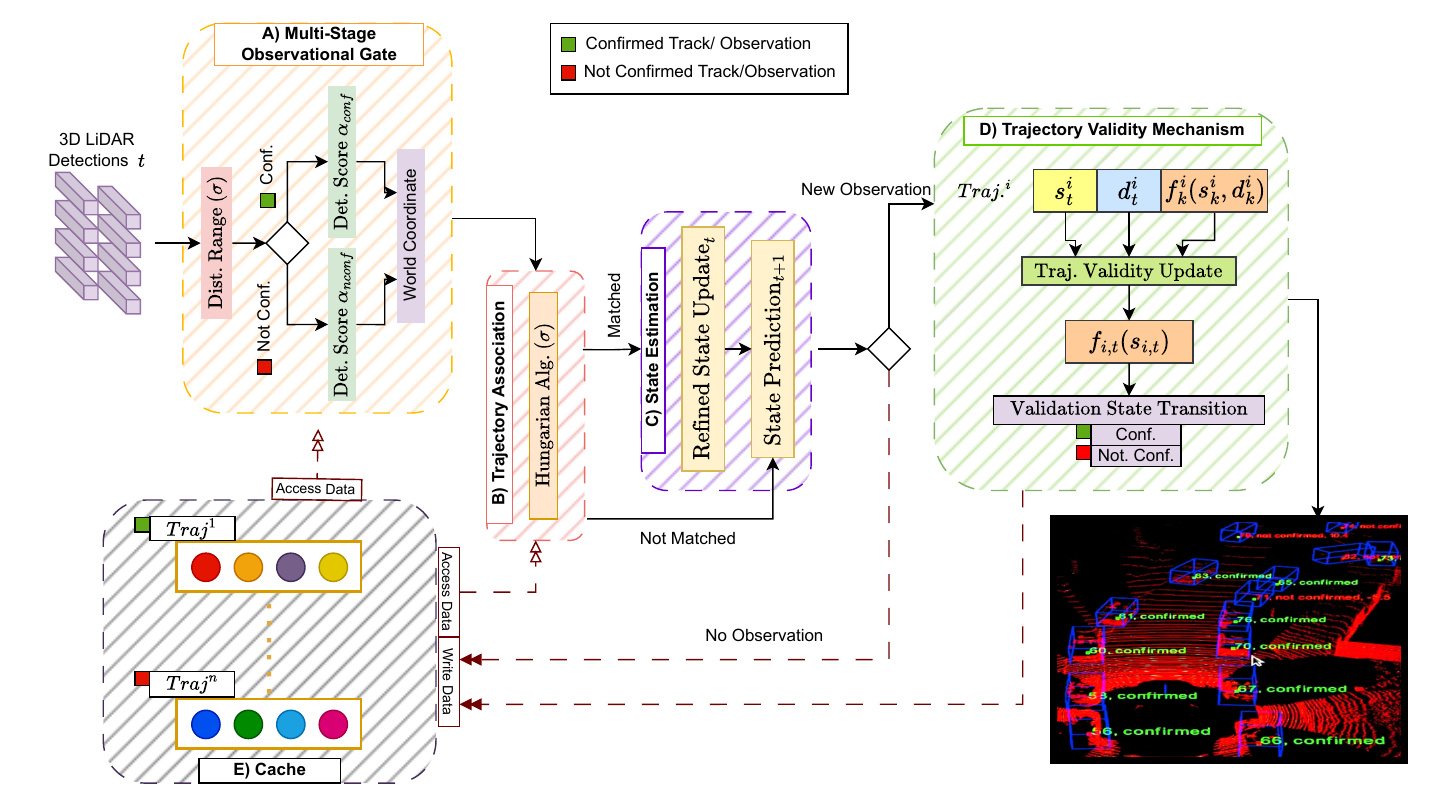}
  \caption{An overview diagram of the proposed method (RobMOT) that consists of five modules. \textcolor{orange}{\textbf{A) Multi-stage observational gate}} that prevents detections from overloading the framework while allowing detections that potentially belong to confirmed trajectories to pass. \textcolor{red}{\textbf{B) Trajectory association}} that associates trajectories stored in the cache with current detections. \textcolor{violet}{\textbf{C) State estimation}} is used to update and predict the following spatial state of trajectories after the association stage while mitigating the spatial localization noise. \textcolor{darkgreen}{\textbf{D) Trajectory validity mechanism}} that updates the temporal validity score for trajectories associated with new observations. \textbf{\textcolor{black}{E) Cache}} that maintains active trajectories while terminating trajectories with high uncertainty about their estimated spatial state.}
  \label{fig:pipeline}
\end{figure*}

3D MOT methods on LiDAR point can be categorized into learning and non-learning (classical) approaches. Part of learning-based methods in the literature involves end-to-end deep learning models that combine object detection and tracking~\cite{track_det_deep_1, track_det_deep_2, track_det_deep_3, track_det_deep_4, 9736652}. Meanwhile, other methods adapt tracking-by-detection paradigm~\cite{track_deep_1, track_deep_2, track_deep_3, track_deep_4}  when the detection of objects is given and the aim is to perform temporal tracking of the objects. The recent research integrates attention mechanism and transformers concepts~\cite{10208977, Mancusi_2023_ICCV} into MOT to enhance the tracking performance. However, learning-based solutions require evolved hardware to be applicable to real-time applications.\\
Meanwhile, classical MOT solutions~\cite{intro_false_positive, Rethink_mot, deepfusion_mot, intro_tra_by_det_kim_eager, oc_sort, nagy, CasTrack, stadler2023improved,zhang2022bytetrack} have gained much attention due to their low latency and the comparative performance with learning-based solutions. 
Despite recent advancements in the classical methods, challenges such as the limitations of the KF, ghost tracks, and memory management remain, particularly in autonomous driving applications where the environment constantly changes. This section will shed light on these challenges and discuss solutions adopted in the literature. \\
\textbf{Kalman filter} is widely used in tracking-by-detection methods~\cite{deepfusion_mot, intro_tra_by_det_kim_eager, nagy, CasTrack} for autonomous driving because of its performance in filtering noise from state estimation while dealing with noisy measurements. If the noise is not properly modeled in the filter, it can cause deviation in the state estimation, as the filter's estimation is sensitive to imprecise measurements. Cao et al.\cite{oc_sort} prevent error accumulation in state prediction caused by periodic occlusion of objects by refining the object's trajectory after object reappearance. However, they need to recover the object to refine its trajectory. Recent work in vehicle tracking\cite{zhou2021short} has shown that integrating probabilistic models and driver preview characteristics can improve state estimation and prediction, offering potential insights for enhancing Kalman filter-based tracking, especially in scenarios involving occlusion. Due to the point cloud's sparsity, our research shows that detections from deep learning detectors on the 3D LiDAR point cloud suffer from spatial distortion around the actual location of the objects (Section~\ref{subsec:trajectory_ass_state_estm}). This spatial distortion results in drift in state estimation for occluded objects as their motion parameters estimation becomes imprecise. \\ \textbf{Ghost tracks}  are false positive detections generating illusion tracks that could eventually lead to wrong navigation of the autonomous vehicle (AV). Since the tracking-by-detection paradigm relies heavily on detection accuracy, ghost tracks in object tracking can significantly harm the tracking performance. Kim et al. \cite{intro_tra_by_det_kim_eager} introduce a multi-stage data association method to overcome this issue by employing multiple detection sources. Similarly, Zhang et al.~\cite{zhang2022bytetrack} employs a two-stage association by prioritizing the association of detections with a high confidence score in the first stage and the second stage for a low confidence score to prevent potential ghost tracks from being associated with valid tracks. Nevertheless, they still depend on predefined thresholds to prevent false positives. The downside of a fine-tuned threshold while filtering ghost tracks is the potential of leaking false positive detections that exceed the threshold while blocking detections that belong to actual tracks. Other approaches~\cite{Rethink_mot,deepfusion_mot,stadler2023improved} incorporate a filtration step to address false positive detections. 
Wang.X  et al.~\cite{deepfusion_mot} 
confirm the legitimacy of a track when three consecutive detections of that track have been observed. In contrast, Wang.L  et al. ~\cite{Rethink_mot} select objects for the highest 30\% detections in the confidence score. However, the assumptions provided by these solutions may not apply to all scenarios, particularly distant objects or objects subject to periodic occlusions.\\
Stadler and Beyerer~\cite{stadler2023improved}, on the other hand, attempt to identify ghost tracks and filter them out by excluding bounding boxes that have high intersection over union (IoU) with others; however, this approach has two limitations; the first drawback is that it can exclude non-ghost track detections in congestion scenarios while it assumes ghost tracks are always generated from duplicated detections for the same object even though Figure~\ref{fig:ghost_real_comp} shows a ghost track (Bounding box with ID:50) that appears far from other objects, as a false detection from the detector.\\
\textbf{Memory management} plays a critical role in addressing false positive detections and object occlusion. It is most common in MOT methods~\cite{deepfusion_mot, intro_tra_by_det_kim_eager, nagy, CasTrack} to extend the lifetime of tracks in the memory to handle occlusions, particularly prolonged ones. However, this expansion comes with the cost of late elimination for ghost tracks that harm the tracking performance.\\
Wu et al.~\cite{CasTrack} resolve the issue by associating a confidence score to pre-observed objects in object association. This score decreases over time when the object becomes not observed for a period of time to prevent potential associations with false positive detections. Nonetheless, this approach cannot recover objects under prolonged occlusion as their association score will decrease.\\
\noindent This work proposes RobMOT, a framework to mitigate state estimation deviation caused by the spatial localization noise from deep learning detectors. It also provides an online adaptive identification mechanism for ghost tracks integrated with a multi-stage observational gate that prevents legitimate track detections from being blocked. The framework handles the trade-off between fast removal for ghost tracks and recovering prolonged occlusions using uncertainty-based memory, which allows quick elimination for ghost tracks while extending the lifetime of legitimate tracks. 
\section{Methodology}
As shown in Figure~\ref{fig:pipeline}, the framework comprises five modules. \textbf{A) Multi-stage observational gate: } prevents overwhelming the framework with unnecessary observations while allowing potential valid observations to pass. \textbf{B) Trajectory association: } associates trajectories in the Cache with current detections. \textbf{C) State estimation:} a refined KF that updates state estimation of trajectories and predicts their next state while mitigating the spatial localization noise of the employed detector. \textbf{D)Trajectory validity mechanism:} update the temporal validity score for trajectories with new observations. \textbf{E) Cache:} holds active trajectories in the system while terminating those with high uncertainty about their spatial state estimation.

\subsection{Multi-Stage Observational Gate}
\label{subsec:Observational_Association_Gate}
Although the trajectory validity mechanism stage (Section~\ref{subsec:Trajectory_Validity}) is responsible for discrimination between ghost and legitimate tracks, detectors on the 3D point cloud provide an enormous number of detections that can significantly impact the run time of the framework. Hence, the multi-stage observational gate is introduced to limit the overflow of the detections and allow detection that potentially belongs to legitimate tracks to pass even with a low detection score.\\
The incoming detections can be categorized into three types: observations that belong to legitimate tracks in the Cache, new tracks, and false observations that lead to the formation of ghost tracks. Initially, a threshold $\alpha_{nconf}$  is assigned to prevent overloading the MOT method by detections; however, this threshold is assigned a minimal detection score value that prevents the detections overflow, regardless of false observations, since ghost tracks are handled later by the trajectory validity mechanism.

Although assigning $\alpha_{nconf}$ with a small value allows a high volume of observations to pass through the system, it can block observations belonging to legitimate tracks in the memory. Hence, observations that potentially belong to legitimate tracks in the cache are identified using the same Euclidean distance threshold $\sigma$ used in the association step (see Section~\ref{subsec:trajectory_ass_state_estm}).Observations with a distance less than $\sigma$  to a legitimate track are identified as potential legitimate observations.\\
Although potential legitimate observations can pass directly to the system, they still cause high latency because of flooding detections from deep learning models on the point cloud. Thus, an additional detection $\alpha_{conf}$ score filtration is provided such that $\alpha_{conf}\leq\alpha_{nconf}$ to allow legitimate observations to enter the system while preventing overloading the system. $\alpha_{nconf}$ and  $\alpha_{conf}$ are fine-tuned so that no enhancement in the tracking performance is observed since their preliminary purpose is to prevent overloading the system. Lastly, the filtered observations are projected from the LiDAR coordinate to the world coordinate.\\

To summarize, the multi-stage observational gate identifies potential legitimate detections, donated by $n$, from the utilized detector using legitimate trajectories stored in the memory, donated by $l$, that contains $m$ trajectories such that $l \leq m$. In the worst case, this module's run-time complexity is $O(nl)$. Algorithm~\ref{alg:observational} demonstrates these procedures in detail.
\begin{algorithm}[!tbh]
\small
    \caption{Multi-Stage Observational Gate}
    \label{alg:observational}
    \begin{algorithmic}[0]
        \STATE \textbf{function} MultiStageObservationalGate($D, T, \sigma, \alpha_c, \alpha_n$)
          \STATE \textbf{Input:}
        \STATE \quad $D$: List of 3D detections from LiDAR
        \STATE \quad $T$: List of trajectories in the cache
        \STATE \quad $\sigma$: Euclidean distance threshold
        \STATE \quad $\alpha_{conf}$: Detection score threshold for potential confirmed observations
        \STATE \quad $\alpha_{nconf}$: Detection score threshold for non-confirmed observations
        \STATE \textbf{Output:}
        \STATE $P$: List of detections that passed the gate
        \newline
        \STATE $P \leftarrow$ []
        \FOR{each $det$ in $D$}
            \IF{$\text{$det$.score} \leq \alpha_{conf}$}
                \STATE continue
            \ENDIF
            \STATE $det.confirmed \leftarrow false$
            \IF{\text{$det$.score} $<\alpha_{nconf}$}
            
            \FOR{each $traj$ in $T$.getConfirmed()}
                \STATE $dist \leftarrow \text{EuclideanDistance}(\text{$det$.center}, \text{$traj$.lastStateEst.center})$
                \IF{$dist \leq \sigma$}
                    \STATE $det.confirmed \leftarrow true$
                    \STATE break
                \ENDIF
            \ENDFOR
            \ENDIF
            \IF{$det.confirmed$ \OR $\text{$det$.score} \geq \alpha_{nconf}$}
                \STATE $detW \leftarrow$ FromLidarToWorld($det$)
                \STATE Add $detW$ to $P$
            \ENDIF
        \ENDFOR
        \RETURN $P$
        \STATE \textbf{end function}
    \end{algorithmic}
\end{algorithm}
\subsection{Trajectory Association and State Estimation with a Refined Kalman Filter for Noise Mitigation}
\label{subsec:trajectory_ass_state_estm}
In the trajectory association stage, the permitted observations are associated with trajectories cached in the framework. The Hungarian algorithm~\cite{kuhn1955hungarian} is used with the algebraic association strategy followed in~\cite{nagy}. The association for an observation is performed by selecting a trajectory with the shortest Euclidean distance between the estimated state and the observation. When the shortest distance to an observation exceeds $\sigma$, a new trajectory is created for the observation. \\
In the next stage (State estimation), the framework performs next-state estimation for all trajectories cached in the system. Trajectories associated with an observation will pass through state update and state prediction as demonstrated in Figure~\ref{fig:pipeline}. Other trajectories will go directly through state prediction. The framework utilizes KF with a constant acceleration motion model so that the state space of a trajectory consists of its spatial location, velocity, and acceleration in $x$ and $y$ directions. 
\begin{figure}[!tbh]
  \centering
  \includegraphics[width=\linewidth]{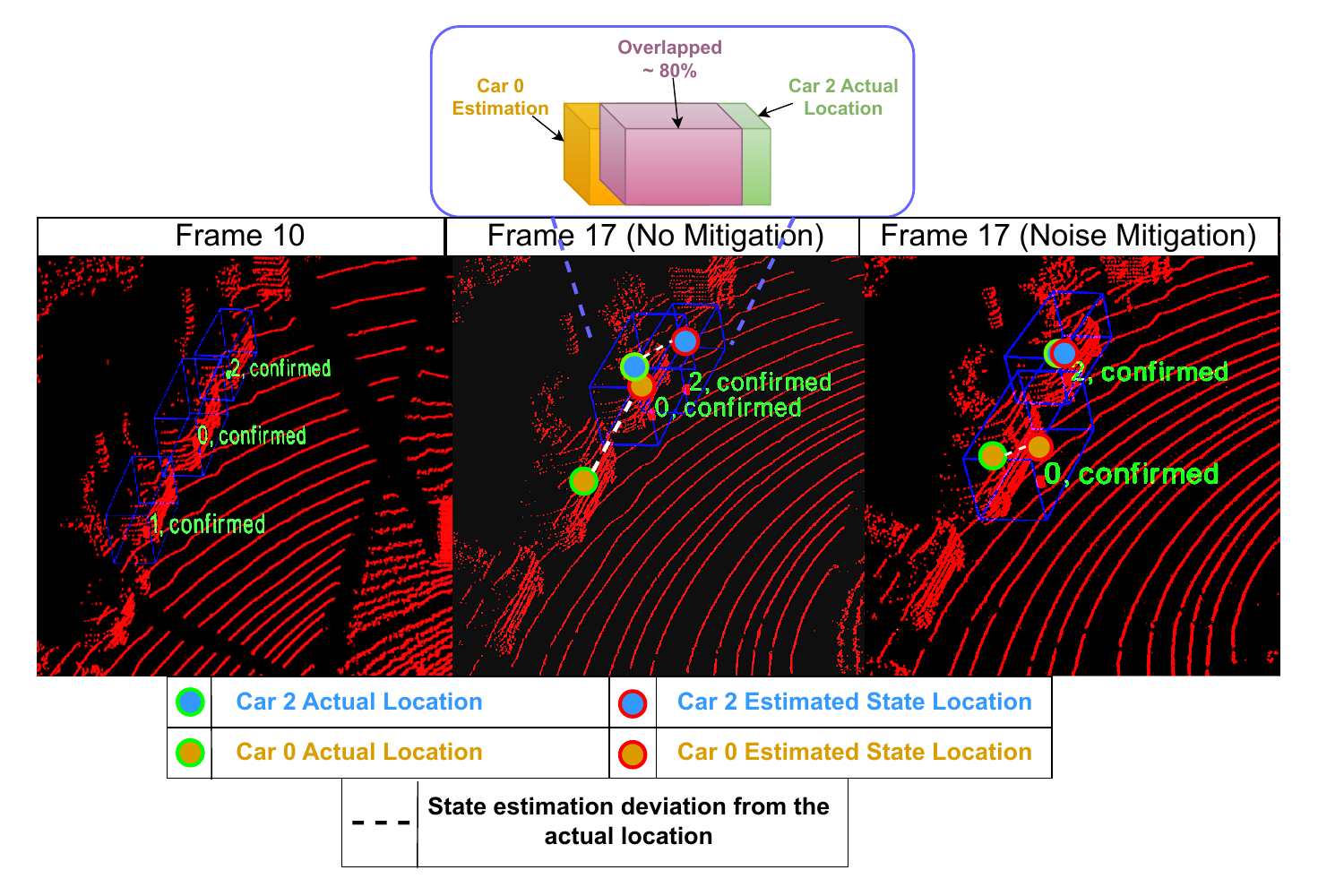}
  \caption{A scenario demonstrates the impact of detection noise on parked cars. Initially, the state estimation of the three cars is positioned correctly in frame 10. The middle frame, frame 17, shows drift in state estimation of cars with ID 0, \textbf{\textcolor{orange}{orange circle}},  and 2,  \textbf{\textcolor{blue}{blue circle}}, caused by the noise. The state deviation from the original position of the cars is marked by \textbf{disconnected white line}. Considering the proposed noise mitigation  (the third scene on the left), the state estimation of cars with ID 2 has no deviation, while the Car 0 deviation has been reduced significantly.}
  \label{fig:traj_drift_our}
\end{figure}

It has been observed that state estimation drifts for objects encountering occlusion or leaving the sensor field of view (off-scene). This phenomenon is also observed in the literature, as displayed in Figure~\ref{fig:trajectory_drift_scene} when the estimated trajectory of a parked car with ID 4 drifts away from the expected location during the occlusion. Figure~\ref{fig:traj_drift_our} shows another scenario when two parked cars with ID 0 and 2 drift from the actual position after being off-scene. Our investigation of the phenomenon concludes that this deviation is caused by trembling in detection bounding boxes around objects' locations, which results in illusion movement for these objects that eventually impacts state update and prediction in KF. \\
This noise is modeled using the algebraic deviation, shown in  Equation~\ref{eq:process_noise_eq}, between the detected bounding boxes $z^{det}_{ij}=(x_{ij}, y_{ij})$ and the actual location of the objects $z^{gt}_{ij}=(x_{ij}, y_{ij})$ across $k$ observations for $N$ objects using KITTI training dataset. 
\begin{equation}
\small
\begin{split}
    \mu_x &= \frac{1}{\sum_{i=1}^N k_i}\sum_{i=1}^N\sum_{j=1}^{k_i} \left(z^{gt}_{ij}(x) - z^{det}_{ij}(x)\right) \\
    \sigma_x^2 &= \frac{1}{\sum_{i=1}^N k_i}\sum_{i=1}^N\sum_{j=1}^{k_i} \left(\left(z^{gt}_{ij}(x) - z^{det}_{ij}(x)\right) - \mu_x\right)^2
\end{split}
\label{eq:process_noise_eq}
\end{equation}

\begin{figure}[tbh]
  \centering
  \includegraphics[width=7cm, height=3.5cm]{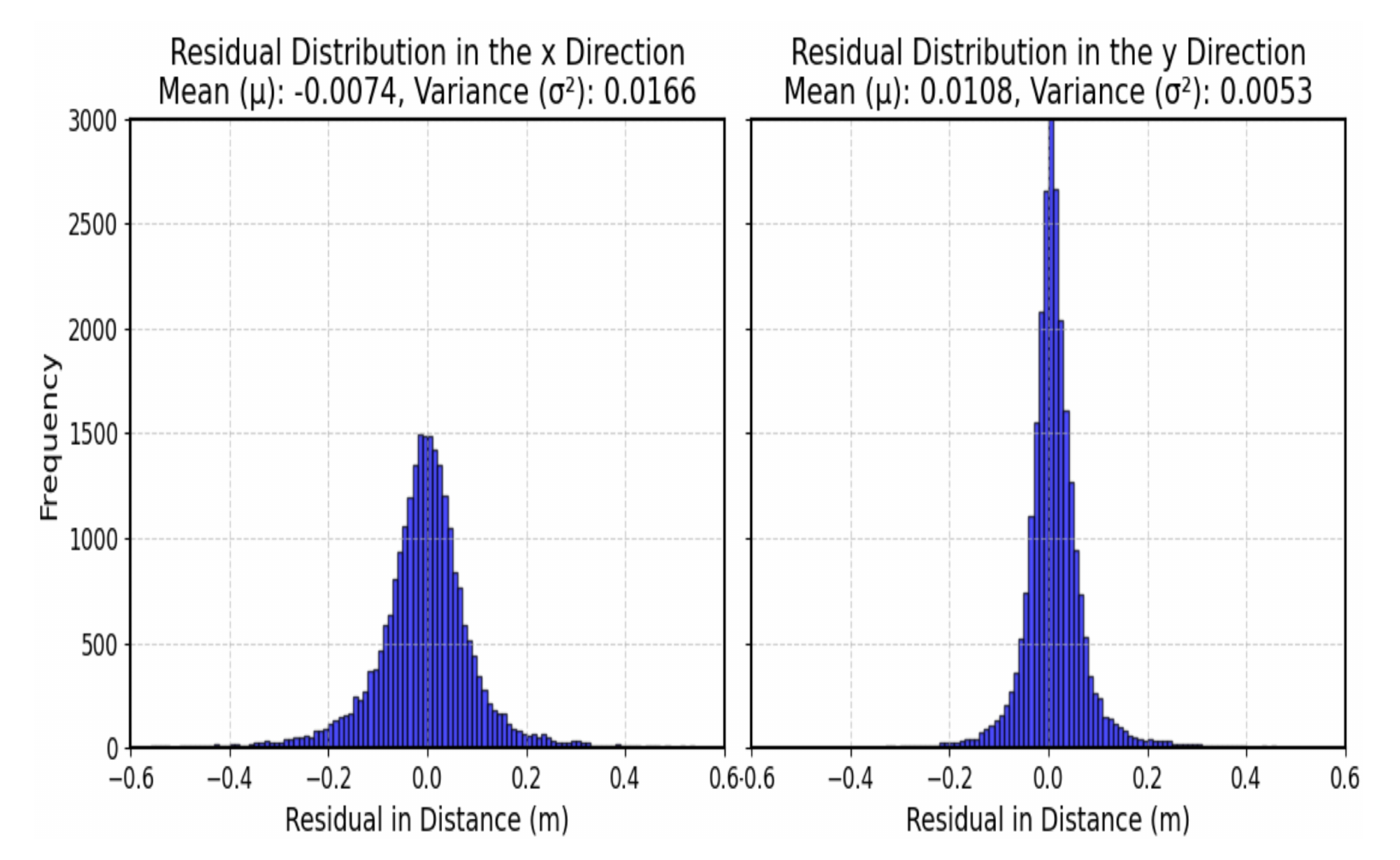}
  \caption{The distribution of the fluctuation noise associated with VirConv~\cite{VirConvDet_CVPR_23} detections. The distribution shows Gaussian noise across five different detectors~\cite{VirConvDet_CVPR_23, casA_detector, second_detector, point_rcnn_detector, pv-rcnn_detector}.}
  \label{fig:dis_graph}
\end{figure}

As depicted in Figure~\ref{fig:dis_graph}, the error distribution shows a Gaussian shape for the noise of the VirConv detector~\cite{VirConvDet_CVPR_23} across $x$ and $y$ LiDAR coordinates, which facilitates its integration into KF.  Consequently, $2\times2$ covariance matrix $D_t$  is driven that represents the impact of noise on the detections (measurements). In particular, noise varies from one detector to another as to the variance in detection performance; hence, $D_t$ should be recomputed as the employed detector changes. By applying Equation~\ref{eq:process_noise_eq} on $y$ coordinate, a covariance matrix of the noise is obtained, Equation~\ref{eq:covariance_mat}.
\begin{equation}
\small
\mathbf{D_t} = \left[
    \begin{matrix}
        \sigma_x^2 & 0\\
        0 & \sigma_y^2
    \end{matrix}
    \right]
    \label{eq:covariance_mat}
\end{equation}

Since this noise is attached to the incoming detections $z_t$ at \textcolor{darkgreen}{time} $t$,  it harms the computation of the residual term $\hat{y}_t$, which is the difference between the current measurement $z_t$ and state estimation $\hat{x}_{t|t-1}$ predicted at $t-1$. Consequently, the noise impact on the residual term leads to incorrect updates for the motion parameters of the object. The innovation covariance $S_t$ is responsible for weighting the innovation residual $\hat{y}_t$ through the Kalman gain $K_t$. Therefore, the covariance matrix $D_t$ needs to be added to the innovation covariance $S_t$ to compensate for the impact of noise on the updated state. Hence, the updated state estimation $\hat{x}_{t|t}$ will consider the impact of the deviation noise on the observation.\\
\begin{equation}
\small
\begin{aligned}
\text{Innovation Residual: }&\tilde{y}_t = z_t - H_t \hat{x}_{t|t-1} \\
\text{Innovation Covariance: }&S_t = H_t P_{t|t-1} H_t^T + R_t + D_t\\
\text{Kalman Gain: 
}&K_t = P_{t|t-1} H_t^T S_t^{-1} \\
\text{State Estimation Update: }&\hat{x}_{t|t} = \hat{x}_{t|t-1} + K_t \tilde{y}_t \\
\text{Estimation Covariance Update: }&P_{t|t} = (I - K_t H_t) P_{t|t-1}
\end{aligned}
\label{eq:kf}
\end{equation}
The main reason for not using $R_t$ instead is because it models the incorrectness of the measurement for the point cloud from LiDAR with respect to the real world, while $D_t$ models the incorrectness of the detection process with respect to the LiDAR point cloud, which means they are independent. Equation~\ref{eq:kf} demonstrates the updated state of KF after adapting $D_t$ noise. 
$H_t$ is the observation model to map from estimation to measurement space at time $t$, while $P_{t|t-1}$ is the estimation uncertainty at time $t$ given measurement at $t-1$. Section~\ref{subsec:Trajectory_Termination_and_Caching} will describe how $P_{t|t}$  contribute to trajectory termination in the cache. 
\subsection{Trajectory Validity Mechanism}
\label{subsec:Trajectory_Validity}
The objective is to monitor the validity (legitimacy) of the existing trajectories in the system while identifying invalid ones, called ghost tracks. Each track has a certainty score that represents its legitimacy. This score is updated whenever a new observation of the track is available. The certainty score changes based on the likelihood of the trajectory being a ghost track. This mechanism is inspired by humans' ability to perceive and identify unfamiliar objects. With multiple observations, human perception can verify and identify unfamiliar objects. By observing the characteristics of ghost tracks, the ghost tracks are usually formed from intermittent sequences of observations with low to moderate detection scores. Based on these characteristics, the certainty score for track $i$ changes when a new observation is associated with the track at time $t$. The certainty score changes according to the detection score $s_i^t$ and the duration of absence $d^i_t$ since the last observation associated with the track.\\

{\small
\vspace{-0.3cm}
\begin{align*}
& s^i_t : \text{detection score of the current observation associated with track }\\ &i \text{ at time } t \\
& d^i_t : \text{absence duration since the last associated observation at time } k\\ &\text{ to track } i \text{ and the current observation at time } t \\
&k: \text{The time stamp of the last associated observation to a trajectory}\\
& f^i_t(s^i_t, d^i_t) : \text{certainty score of trajectory } i \text{ at time } t \\
& f^i_k(s^i_k, d^i_k) : \text{certainty score of trajectory } i \text{ at time } k\\
\end{align*}
\begin{equation}
\begin{aligned}
f^i_t(s^i_t, d^i_t) &= s^i_t e^{-d^i_t} - \frac{d^i_t}{s^i_t} + f^i_k(s^i_k, d^i_k),\text{ where } s^i_t >0\\
d^i_t &= t - (k + 1)
\end{aligned}
\label{eq:validation_track}
\end{equation}
}
As shown in Equation~\ref{eq:validation_track}, the formulation consists of three terms:  \textbf{Confidence decay} term $s^i_t e^{-d^i_t}$ (reward), \textbf{Absence duration} term (penalty) $\frac{d_t}{s^i_t}$, and \textbf{Temporal validity} $f^i_k(s^i_k, d^k_t)$ term.

\subsubsection{Confidence decay}
The term is responsible for increasing the certainty score of track $i$ as the detection score  $s^i_t$ of the associated observation increases as a reward, indicating the detection model's confidence in this observation. However, ghost tracks with moderate detection scores can receive high rewards, potentially misleading the validation mechanism. Thus, a decay rate,$e^{-d^i_t}$, is attached  to the detection score $s^i_t$. The decay $e^{-d^i_t}$  increases as the duration $d^i_t$ of the current observation at time $t$ to the previous observation at time $k$ increases. Given the intermittent characteristic of ghost tracks, the decay term decreases the detection score for potential ghost tracks in the system.
\subsubsection{Absence duration}
The penalty term $\frac{d^i_t}{s^i_t}$ lowers the certainty score as the absence duration $d^i_t$ of a new observation for the track increases. This term will diminish the impact of the reward with an extended absence duration that indicates a high potential of being a ghost trajectory. However, this term can harm legitimate tracks that encounter occlusion. Since observations of legitimate tracks tend to have high detection scores $s^i_t$, the inverse proportion is included in the detection score to decrease the impact of this term on the legitimate tracks. 
\subsubsection{Temporal validity}
The inclusion of $f^i_k(s^i_k, d^i_k)$ integrates the previous certainty score calculated at time $k$ to the current certainty score $f^i_t(s^i_t, d^i_t)$ of time $t$. This term adds temporal validation for trajectories in the system by incorporating historical certainty into the current certainty score so that  $f^i_t(s^i_t, d^i_t)$ represents a certainty score of all trajectory states.

\begin{figure}[tbh]
  \centering
  \includegraphics[width=7cm]{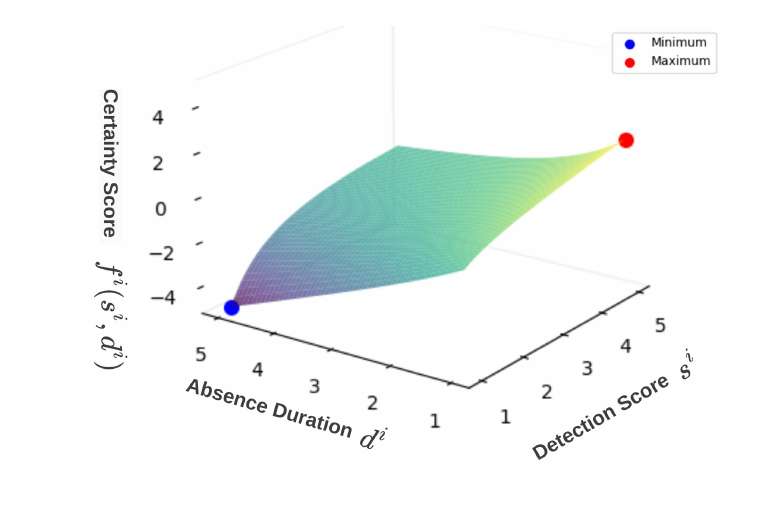}
  \caption{A graphical visualization of the certainty scoring formula over detection scores of range $[0,5]$ and absence duration $[0,5]$ frames. The \textcolor{red}{red point} is the highest certainty score when the detection score is highest while the absence duration is lowest. In contrast, the \textcolor{blue}{blue point} is the lowest certainty score when the detection score is at the lowest value, while the absence duration is the highest. The surface shows the remaining values.}
  \label{fig:form_graph}
\end{figure}
A visual representation of the distribution of Equation~\ref{eq:validation_track} is highlighted in Figure~\ref{fig:form_graph}; the relation between the detection score, the absence duration, and its impact on the certainty score. Even with low detection scores, Equation~\ref{eq:validation_track} returns a higher reward without decay ($e^-{d_t^i} $) or penalty ($\frac{d_t^i}{s_t^i}$) as long as the observation is consistent (too small to zero absence duration). Thus, the trajectory of true positive detections with low detection scores caused by partial occlusion or distance can still be verified.

This mechanism monitors the validity of trajectories in the system individually until a certainty score of a trajectory exceeds a pre-defined threshold $\alpha_{legit}$ that indicates the trajectory is legitimate. In this case, the validation mechanism changes the trajectory status from ``not confirmed'' to ``confirmed'' and terminates the validation mechanism to this trajectory. Thus, confirmed trajectories cannot be changed to ``not confirmed" as the mechanism is terminated for these tracks. 

\subsection{Trajectory Termination and Caching}
\label{subsec:Trajectory_Termination_and_Caching}
\begin{figure}[!tbh]
    \centering
    \includegraphics[width=\linewidth]{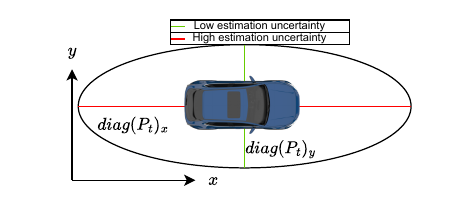}
    \caption{ An illustration of spatial state estimation uncertainty for a car in a 2D dimension shows \textcolor{red}{high estimation uncertainty} along the x-axis while \textcolor{darkgreen}{low uncertainty} on the y-axis.}
    \label{fig:estimation_uncertain}
\end{figure}

The cache holds all active trajectories. In the literature, a trajectory is terminated or discarded when there are no further observations for that trajectory for several frames. Generally, it is not trivial to define an absence threshold, as it could risk eliminating objects under prolonged occlusion or maintaining ghost tracks for a long time. Instead, trajectories are discarded when there is high uncertainty about their estimation, as that uncertain estimation is more likely to be mismatched. Each trajectory has an estimation uncertainty $P_{t|t}$ from Equation~\ref{eq:kf} on $x$ and $y$ direction that forms an ellipse, as shown in Figure~\ref{fig:estimation_uncertain}. The trajectory is discarded when estimation uncertainty ${diag(P_t)_x, diag(P_t)_y}$ becomes high in either direction, exceeding $\sigma_{est\_certainity} = 4$ meters. This strategy serves trajectories under occlusion and fast elimination for ghost tracks. As per our observation, ghost tracks tend to have high growth in estimation uncertainty because of their limited observations, while legitimate tracks maintain low estimation uncertainty, even those under occlusion.

\section{Results and Discussion}
\label{sec:result}
The framework is evaluated using two benchmark datasets for autonomous cars: KITTI~\cite{Geiger2012CVPR} and Waymo Open Dataset~\cite{9156973} (WOD). The KITTI dataset consists of $50$ sequences, split into $21$ training sequences and $29$ testing sequences recorded around Karlsruhe, Germany, that involve a variety of challenging conditions. Meanwhile, WOD contains about $1150$ segments recorded in various locations across the United States.\\
The evaluation of the proposed approach consists of an extensive quantitative and qualitative evaluation with recent benchmarks in this section and a sensitive analysis of each proposed component to evaluate its contribution to the tracking performance, which will be covered in the ablation study, Section~\ref{sub:ablation_study}.
\subsection{Implementation Details}
\label{sec:implementation}
This research is implemented in C++17. The modules have been implemented from scratch to ensure the efficiency and precise implementation of the proposed methods. The Eigne3 library is used to perform the algebraic procedures discussed in this work. Meanwhile, OpenCV 4.6 and PCL 1.14 are used to visualize the result. The tracking evaluation of KITTI validation and training datasets was done using their official evaluation tool~\cite{luiten2020trackeval}. The experiments are conducted on a machine with a processor \textit{AMD Ryzen 9} and 32 GB RAM; no GPU is involved since the framework utilizes a single CPU.

\subsection{Evaluation with Benchmarks}
\label{sub:evaluation_with_benchmarks}
This section consists of quantitative tracking performance evaluation with the latest state-of-the-art MOT methods on KITTI and WOD datasets and qualitative evaluation focusing on challenging occlusion conditions with the top-performing benchmark selected from the quantitative evaluation.\\

\begin{figure*}[!b]
  \centering
  \includegraphics[width=\linewidth]{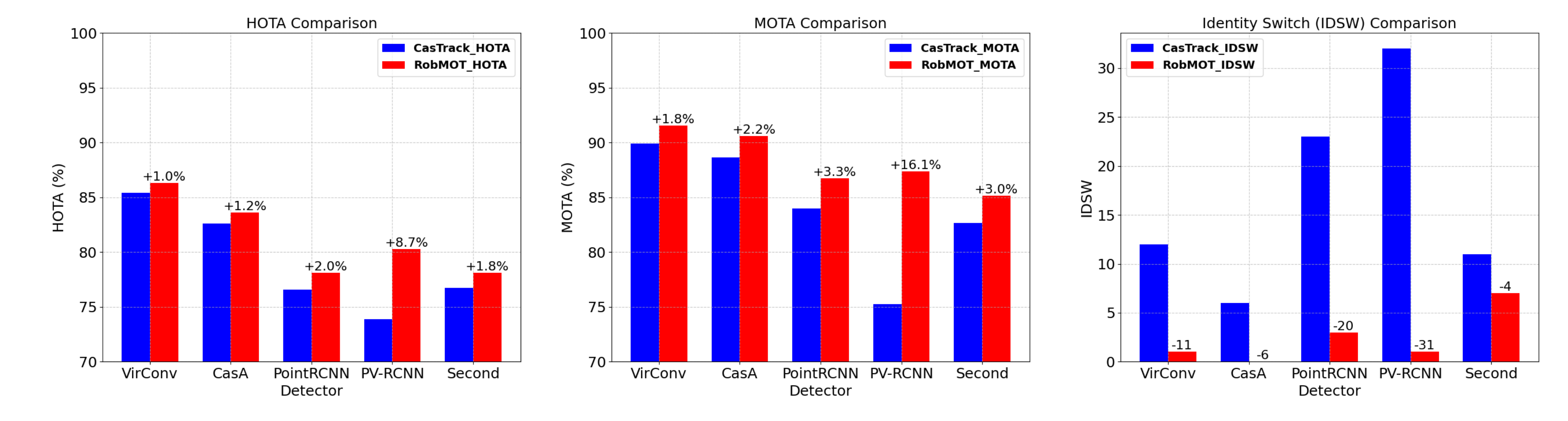}
\caption{Tracking performance comparison on KITTI dataset \textit{validation split} with CasTrack~\cite{CasTrack} across five different detectors: VirConv~\cite{VirConvDet_CVPR_23}, CasA~\cite{casA_detector}, PointRCNN~\cite{point_rcnn_detector}, PV-RCNN~\cite{pv-rcnn_detector}, and Second~\cite{second_detector}. RobMOT (in \textbf{\textcolor{red}{red}}) consistently outperforms CasTrack (in \textbf{\textcolor{blue}{blue}}) across all detectors, with significant improvements in both HOTA and MOTA metrics, particularly with the PV-RCNN detector, and a substantial reduction in IDSW. Additionally, RobMOT demonstrates a minimum 40\% increase in processing speed (fps)  across all detectors, highlighting its efficiency and robustness. The percentage improvements are indicated in \textbf{\textcolor{darkgreen}{green}}.}
  \label{fig:hota_mota_comp}
\end{figure*}

The primary metrics used in the evaluation are HOTA and MOTA (See Equation~\ref{eq:hota_mota}). MOTA metric relies mainly on the false positives (FP) (ghost tracks), false negatives (FN) (missed tracks), and switch in tracks' identity (IDSW). Since a single detector is unified among experiments, FN remains the same as long as the detector is not changed so that the enhancement in MOTA comes from fewer ghost tracks (FP) and IDSW. On the other hand, HOTA is concerned about detection accuracy (DetA) and association accuracy (AssA). AssA represents the average localization of the state estimation and the actual object location based on how well the state estimation bounding box overlaps with the actual object location ($IoU_{matched}$). Accordingly, the improvement in HOTA while unifying the detector indicates improvement in the state estimation localization.

\begin{equation}
\small
   \begin{split}
\text{MOTA} &= 1 - \frac{\sum_{t} (FN_t + FP_t + IDSW_t)}{\sum_{t} GT_t}\\
    \text{HOTA} &= \sqrt{\text{DetA} \times \text{AssA}}\\
    \text{AssA} &= \frac{\sum \text{IoU}_{\text{matched}}}{\text{Total correct matches}}
\end{split}
\label{eq:hota_mota}
\end{equation}
  
To this end, MOTA can measure the contribution of the proposed trajectory validity mechanism for ghost tracks. At the same time, HOTA can indicate the impact of the proposed noise mitigation term $D_t$ on enhancing state estimation localization for objects in the KF.

\subsubsection{Quantitative Evaluation} 
\begin{table}[htb]
\centering
\caption{\label{tab:kitt_test_eval}The table compares the latest benchmarks on the testing split in the KITTI leaderboard. The  \textbf{highest performance} among classical methods at each metric is highlighted in \textbf{\textcolor{red}{red}}, the  \textbf{second highest in \textcolor{blue}{blue}}, the \textbf{third highest in \textcolor{darkgreen}{green}}, and the \textbf{fourth highest in \textcolor{violet}{violet}}. Methods that use the CasA~\cite{casA_detector} detector are marked by ($*$), while ($\dot{+}$) is used for methods that use the VirConv~\cite{VirConvDet_CVPR_23} detector. }
\label{ch6:tab:robmot_benchmark_eval_kitti}
\begin{adjustbox}{max width=\columnwidth}
\begin{tabular}{@{}|l|c|c|c|c|c|@{}}
\toprule
\textbf{Method} & \textbf{HOTA} & \textbf{MOTA} & \textbf{AssA} & \textbf{AssRe} & \textbf{IDSW} \\ 
\midrule
TripletTrack~\cite{marinello2022triplettrack} & 73.58\% & 84.32\% & 74.66\% & 77.3\% & 322 \\ 
PolarMOT~\cite{10.1007/978-3-031-20047-2_3} & 75.2\% & 85.1\% & 76.95\% & 80.0\% & 462 \\ 
DeepFusion-MOT~\cite{deepfusion_mot} & 75.5\% & 84.6\% & 80.1\% & 82.6\% & \textbf{\textcolor{violet}{84}} \\ 
Mono-3D-KF~\cite{9626850} & 75.5\% & 88.5\% & 77.6\% & 80.2\% & 162 \\ 
PC3T~\cite{9352500} & 77.8\% & 88.8\% & 81.6\% & 84.8\% & 225 \\ 
MSA-MOT~\cite{zhu2022msa} & 78.5\% & 88.0\% & \textbf{\textcolor{violet}{82.6\%}} & 85.2\% & 91 \\ 
\textcolor{darkgreen}{UG3DMOT*~\cite{he20243d}} & 78.6\%&87.98\%&82.28\%&85.36\%&\textbf{\textcolor{blue}{30}}\\
CasTrack*~\cite{CasTrack} & 77.3\% & 86.29\% & 80.29\% & 83.12\% & 184\\ 
VirConvTrack$\dot{+}$~\cite{Wu2023Supplemental, CasTrack} & \textbf{\textcolor{violet}{79.9\%}}& \textbf{\textcolor{violet}{89.1\%}}& \textbf{\textcolor{violet}{82.6\%}}& \textbf{\textcolor{violet}{85.6\%}}&201\\
PC-TCNN~\cite{ijcai2021p161} & \textbf{\textcolor{darkgreen}{80.9\%}} & \textbf{\textcolor{red}{91.7\%}} & \textbf{\textcolor{darkgreen}{84.1\%}} & \textbf{\textcolor{darkgreen}{87.5\%}} &  \textbf{\textcolor{darkgreen}{37}} \\ 
\midrule
RobMOT* (Ours) & \textcolor{blue}{\textbf{81.22\%}} & \textbf{\textcolor{darkgreen}{90.5\%}} &  \textbf{\textcolor{red}{85.8\%}} &  \textbf{\textcolor{red}{89.7\%}} &  \textbf{\textcolor{red}{7}} \\ 
RobMOT$\dot{+}$ (Ours) & \textcolor{red}{\textbf{81.76\%}} &  \textbf{\textcolor{blue}{91.02\%}} &  \textbf{\textcolor{blue}{85.6\%}} & \textbf{\textcolor{blue}{89.25\%}} &  \textbf{\textcolor{red}{7}} \\ 
\bottomrule
\end{tabular}
\end{adjustbox}
\end{table}

The proposed method is evaluated quantitatively on the KITTI testing dataset with the latest published state-of-the-art tracking methods listed in the KITTI leaderboard~\cite{Geiger2012CVPR}. Since the proposed method is an online method (real-time) while the leaderboard combines online and offline methods, Table~\ref{tab:kitt_test_eval} reports the online methods (classical and deep learning methods). The framework outperforms recent online benchmarks in almost all metrics, including the recent deep learning approach PC-TCNN~\cite{ijcai2021p161}, using either CasA~\cite{casA_detector} or Virconv~\cite{VirConvDet_CVPR_23} detector. As our approach is a classical method, the evaluation will concentrate on the recent classical benchmark method~\cite{Wu2023Supplemental, CasTrack} (VirConvTrack/CasTrack) because they achieve the highest performance as per Table~\ref{tab:kitt_test_eval} and the KITTI leaderboard~\cite{Geiger2012CVPR}.\\
RobMOT outperforms CasTrack and VirConvTrack in all metrics using the same detectors, with a margin reaching $3.92\%$ in HOTA, $5.79\%$ in AssA, and $6.85\%$ in association recall (AssRe) with the CasA detector. Notably, the proposed method shows a significant reduction in IDSW in both detectors, which shows its tracking stability.\\

A more intensive evaluation study is conducted on the KITTI evaluation dataset across five different detectors with the method~\cite{CasTrack, Wu2023Supplemental} as it achieves the highest tracking performance among the classical methods. Figure~\ref{fig:hota_mota_comp} represents the outcome of this study in HOTA, MOTA, and IDSW metrics. The proposed method consistently outperforms the benchmark~\cite{CasTrack, Wu2023Supplemental} in all the metrics across the detectors. The performance margin in HOTA reaches $8.7\%$ and $16.1\%$ in MOTA with PV-RCNN detector~\cite{pv-rcnn_detector}. Moreover, our method shows a significant reduction in IDSW, as shown in Figure~\ref{fig:hota_mota_comp} (the third graph), which supports results in Table~\ref{tab:kitt_test_eval} on the test dataset. This significant performance drop in work~\cite{CasTrack} shows the dependency of the method on the detector performance. 

\begin{table*}[!tbh]
\caption{Tracking performance on Waymo Open Dataset testing split with the latest online MOT benchmarks. The color code used in the table to highlight the first-highest (\textbf{\textcolor{red}{red}}) and second-highest (\textbf{\textcolor{darkgreen}{green}}) performance for \textbf{benchmarks use the same detector}. Benchmarks with the CasA~\cite{casA_detector} detector are marked by (*) while the CTRL~\cite{ctrl_detector} detector is marked by (**).}
\centering
\begin{tabular}{@{}|l|l|ll|ll|@{}}
     \toprule
     &&\multicolumn{2}{c|}{Level 1}&\multicolumn{2}{c|}{Level 2}\\
     \midrule
     Method&Range($m$)&MOTA&Mismatch$\downarrow$&MOTA$\uparrow$&Mismatch$\downarrow$\\
    \midrule
    SimpleTrack~\cite{Pang2021SimpleTrackUA}&$[0,30)$&86.23\%&0.02\%&85.42\%&0.02\%\\
    ImmotralTrack~\cite{wang2021immortal}&$[0,30)$ &86.35\%&0.00\%&85.55\%&0.00\%\\
    FastPoly~\cite{li2024fast}*&$[0,30)$&87.89\%&0.01\%&87.07\%&0.01\%\\
    CasTrack~\cite{CasTrack}*&$[0,30)$&88.05\%&0.02\%&87.24\%&0.02\%\\
    CasTrack~\cite{CasTrack}**&$[0,30)$&91.98\%&0.01\%&91.37\%&0.01\%\\
    \midrule
    RobMOT* (Ours)&$[0,30)$&\textcolor{darkgreen}{\textbf{88.10\%}}\textcolor{blue}{(+0.05\%)}&\textcolor{darkgreen}{\textbf{0.01\%}}\textcolor{blue}{(--0.01\%)}&\textcolor{darkgreen}{\textbf{87.29\%}}\textcolor{blue}{(+0.05\%)}&\textcolor{darkgreen}{\textbf{0.01\%}}\textcolor{blue}{(-0.01\%)}\\
    
     RobMOT** (Ours)&$[0,30)$&\textcolor{red}{\textbf{92.17\%}}\textcolor{blue}{(+0.19\%)}&\textcolor{red}{\textbf{0.00\%}}\textcolor{blue}{(-0.01\%)}&\textcolor{red}{\textbf{91.59\%}}\textcolor{blue}{(+0.22\%)}&\textcolor{red}{\textbf{0.00\%}}\textcolor{blue}{(-0.01\%)}\\
    \midrule
    \midrule
    SimpleTrack~\cite{Pang2021SimpleTrackUA}&$[30,50)$&64.09\%&0.08\%&61.14\%&0.08\%\\
    ImmotralTrack~\cite{wang2021immortal}&$[30,50)$&64.41\%&0.01\%&61.45\%&0.01\%\\
    FastPoly~\cite{li2024fast}*&$[30,50)$&66.72\%&0.05\%&63.46\%&0.05\%\\
    CasTrack~\cite{CasTrack}*&$[30,50)$&67.42\%&0.06\%&64.38\%&0.06\%\\
    CasTrack~\cite{CasTrack}**&$[30,50)$&78.07\%&0.08\%&75.08\%&0.08\%\\
    \midrule
    RobMOT* (Ours)&$[30,50)$&\textcolor{darkgreen}{\textbf{67.72\%}}\textcolor{blue}{(+0.30\%)}&\textcolor{darkgreen}{\textbf{0.02\%}}\textcolor{blue}{(+0.04\%)}&\textcolor{darkgreen}{\textbf{64.63\%}}\textcolor{blue}{(+0.25\%)}&\textcolor{darkgreen}{\textbf{0.02\%}}\textcolor{blue}{(-0.04\%)}\\
    RobMOT** (Ours)&$[30,50)$&\textcolor{red}{\textbf{78.78\%}}\textcolor{blue}{(+0.71\%)}&\textcolor{red}{\textbf{0.03\%}}\textcolor{blue}{(-0.05\%)}&\textcolor{red}{\textbf{75.99\%}}\textcolor{blue}{(+0.91\%)}&\textcolor{red}{\textbf{0.02\%}}\textcolor{blue}{(-0.06\%)}\\
    \midrule
    \midrule
    SimpleTrack~\cite{Pang2021SimpleTrackUA}&$[50,+inf)$&36.28\%&0.14\%&32.71\%&0.12\%\\
    ImmotralTrack~\cite{wang2021immortal}&$[50,+inf)$&36.59\%&0.02\%&33.02\%&0.02\%\\
    FastPoly~\cite{li2024fast}*&$[50,+inf)$&40.99\%&0.10\%&37.18\%&0.09\%\\
    CasTrack~\cite{CasTrack}*&$[50,+inf)$&41.81\%&0.08\%&37.90\%&0.07\%\\
    CasTrack~\cite{CasTrack}**&$[50,+inf)$&58.47\%&0.14\%&53.37\%&0.12\%\\
    \midrule
    RobMOT* (Ours)&$[50,+inf)$&\textcolor{darkgreen}{\textbf{42.39\%}}\textcolor{blue}{(+0.58\%)}&\textcolor{darkgreen}{\textbf{0.03\%}}\textcolor{blue}{(-0.05\%)}&\textcolor{darkgreen}{\textbf{38.37\%}}\textcolor{blue}{(+0.47\%)}&\textcolor{darkgreen}{\textbf{0.03\%}}\textcolor{blue}{(-0.04\%)}\\
    RobMOT** (Ours)&$[50,+inf)$&\textcolor{red}{\textbf{59.74\%}}~\textcolor{blue}{(+1.27\%)}&\textcolor{red}{\textbf{0.04\%}}\textcolor{blue}{(-0.10\%)}&\textcolor{red}{\textbf{55.14\%}}\textcolor{blue}{(+1.77\%)}&\textcolor{red}{\textbf{0.03\%}}\textcolor{blue}{(-0.09\%)}\\
     \bottomrule
\end{tabular}
\label{tab:wod_eval}
\end{table*}
Another evaluation is conducted on the WOD testing dataset to measure the generalization of the approach. The dataset consists of simple to moderate tracking conditions labeled ``Level 1" and challenging conditions labeled ``Level 2" in Table~\ref{tab:wod_eval}. Table~\ref{tab:wod_eval} represents the tracking performance of the state-of-the-art methods on WOD using MOTA and Mismatch (miss association) metrics. The table consists of tracking evaluations for objects between different distance ranges from the AV: 0 to 30 meters, 30 to 50 meters, and more than 50 meters. \\
The proposed method, RobMOT, achieves the highest tracking performance in both level 1 and level 2 in MOTA and Mismatch. The performance margin with the other benchmarks increases as the tracking distance range increases, reaching $1.77\%$ in MOTA for objects in the range $[50, +inf)$. That shows the method's capability to track distant objects over the recent benchmarks. Moreover, the proposed method consistently outperforms CasTrack~\cite{CasTrack} while unifying the employed detectors (CasA~\cite{CasTrack} and CTRL~\cite{ctrl_detector}) for both methods.\\

In conclusion, the proposed approach excels in tracking stability and accuracy, significantly reducing ID switches. It outperforms recent state-of-the-art methods~\cite{CasTrack, Wu2023Supplemental} on both the KITTI and WOD datasets, demonstrating robust generalization and efficiency. The quantitative results validate the contribution and exceptional performance of the proposed framework.\\

\begin{figure*}[tbh]
  \centering
  \includegraphics[width=\linewidth,height=11cm]{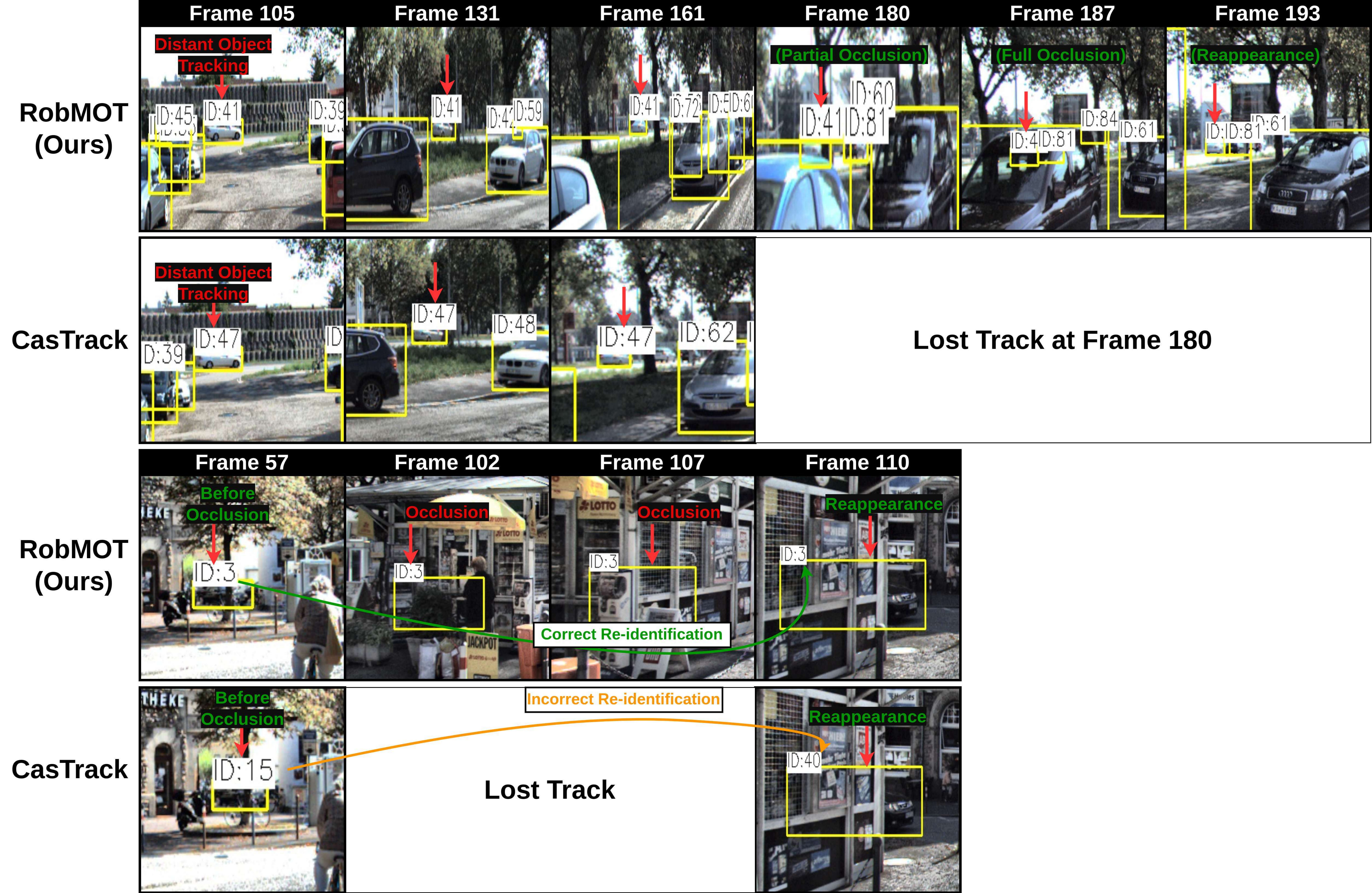}
  \caption{A qualitative evaluation of our framework performance with CasTrack~\cite{CasTrack} in two challenging multi-object tracking scenarios by projecting 3D tracking results into 2D frames. The first two rows show a scenario where a distant car \textit{(ID:41)} faces multiple occlusions and distortion in detection. The second two rows show a scenario of a car \textit{(ID:3)} occluded by a building for about 50 consecutive frames. Our framework tracks the distant car in the first scenario under distortion in the observation and occlusions while providing precise state estimation of the prolonged occluded car in the second scenario with successful recovery after the occlusion.}
  \label{fig:occlusion_comp}
\end{figure*}
\subsubsection{Qualitative Evaluation}
Although the quantitative evaluation shows the overall performance of a tracking approach, it is not sufficient to evaluate the method under challenging tracking conditions such as occlusions. Thus, two challenging conditions are selected to emphasize the contribution of the proposed approach. The first condition represented in the first two rows in Figure~\ref{fig:occlusion_comp} is for a distant car that faces frequent occlusions for about 90 frames. The detector is unified for both methods, ours and CasTrack~\cite{CasTrack}, to measure the contribution of the tracking approach. Our proposed solution maintains tracking for the car with ID $41$ to frame $193$ when the car leaves the field of view of LiDAR, while CasTrack~\cite{CasTrack} lost the car at frame $180$.\\  
Another challenging scenario is illustrated in Figure~\ref{fig:occlusion_comp} in the third and fourth rows. In frame $57$, a distant car behind a tree is occluded by a building in the following frames until frame $110$, when the car partially appears again. Our proposed method shows robustness in state estimation of the car during occlusion, as shown in frames $102$ and $107$, and successfully reassociates the car when it appears in frame $110$. On the other hand, CasTrack~\cite{CasTrack} loses track of the car during the occlusion and considers the re-appearance of the car as a new car associated with a new trajectory. \\
The two scenarios show the robustness of the proposed approach and capability to handle challenging occlusion scenarios compared to the recent state-of-the-art CasTrack~\cite{CasTrack}.\\ 

\subsubsection{Runtime and Speed Comparison}
\begin{figure}[!tbh]
    \centering
    \includegraphics[width=\linewidth]{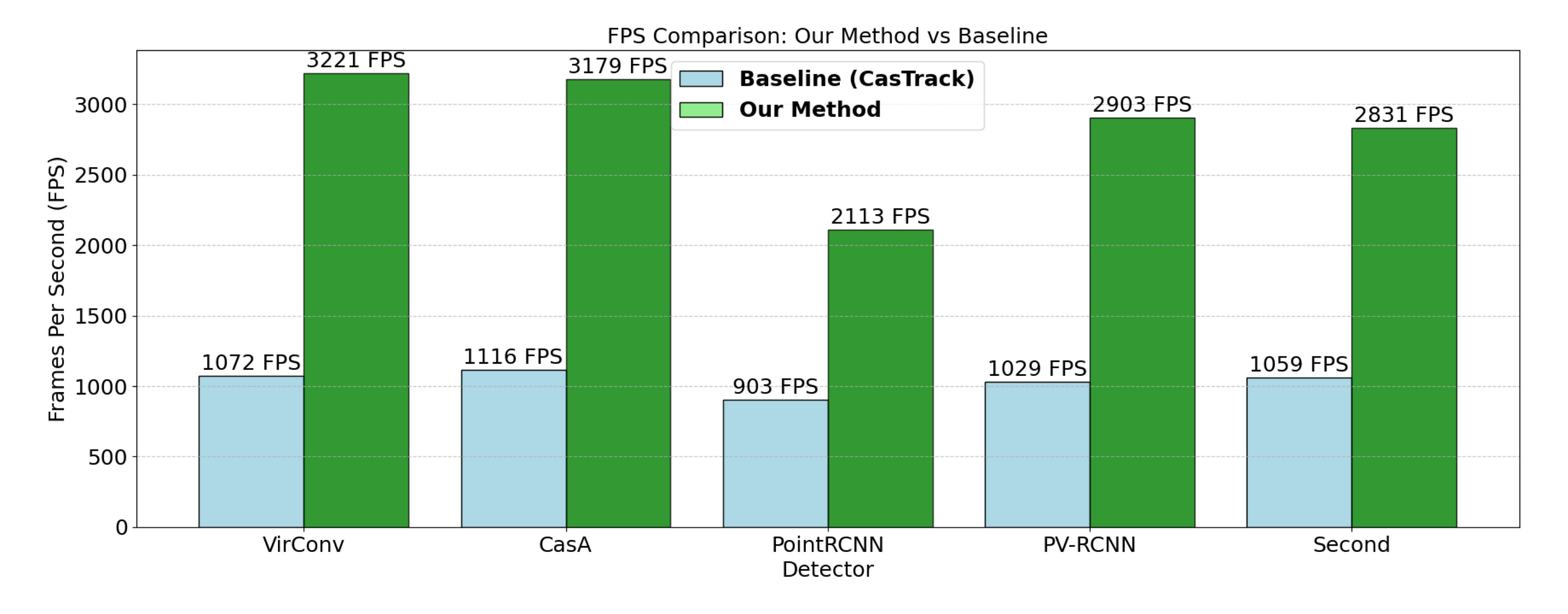}
    \caption{Runtime comparison of processing about 7,500 consecutive frames from KITTI dataset, training and validation splits, across five detectors with benchmark~\cite{CasTrack,Wu2023Supplemental}.}
    \vspace{-0.3cm}
    \label{fig:abl_speed}
\end{figure}

A third study is conducted to provide some insights into the computational efficiency of the proposed approach. Since the number of incoming detections and detections filtration in the MOT method are the main contributors to the run-time efficiency, this experiment compares the performance of the proposed method (RobMOT) with the recent state-of-the-artwork that achieves state-of-the-art tracking speed (Figure 4 in~\cite{CasTrack}) across five different detectors. In this study, the method~\cite{CasTrack, Wu2023Supplemental} is re-run on the same machine as the proposed approach (RobMOT) with hardware stated in Section~\ref{sec:implementation}. Figure~\ref{fig:abl_speed} summarizes the frame per second (FPS) achieved by each method on training and validation KITTI datasets that contains about $7,500$ frames. Our proposed solution significantly reduces computational time compared to method~\cite{CasTrack, Wu2023Supplemental} across the five detectors. Our approach achieves an FPS margin that reaches $2000$ FPS, as shown with the VirConv detector~\cite{VirConvDet_CVPR_23}. This result reflects the applicability of the proposed method for real-time applications, particularly applications with hardware constraints.
\subsection{Ablation Study}
\label{sub:ablation_study}
This section will cover a sensitivity analysis conducted on the proposed method to show the contribution of each component proposed in this work and quantitatively measure its impact on tracking performance. 
\subsubsection{Trajectory Drift}
This research raises concerns about noise injected in the incoming detections, which leads to deviation in the state estimation of objects during occlusion. The noise is observed in multiple scenarios (Figure~\ref{fig:traj_drift_our}) and the recent state-of-the-art method CasTrack~\cite{CasTrack} (Figure~\ref{fig:trajectory_drift_scene}), which emphasizes the existence of this problem. This section demonstrates the quantitative impact of this noise on the tracking performance and the contribution of the proposed noise mitigation approach discussed in this work in improving the performance by mitigating the noise.\\
\begin{table}[htb]
\vspace{-0.2cm}
\centering
\caption{\label{tab:ref_analysis} Performance comparison of different detectors with and without the proposed refinement for noise mitigation $D_t$.}
\begin{adjustbox}{max width=\columnwidth}
\begin{tabular}{|c|c|c|c|c|c|c|}
\hline
\textbf{Metric} & \textbf{$D_t$} & \textbf{VirConv} & \textbf{CasA} & \textbf{PointRCNN} & \textbf{PV-RCNN} & \textbf{Second} \\ 
\hline
\multirow{2}{*}{HOTA} 
& Y & \textbf{85.1\%} & \textbf{82.2\%} & \textbf{76.7\%} & \textbf{78.6\%} & \textbf{77.2\%} \\ 
& N & 82.7\% & 79\% & 73.2\% & 73.8\% & 73.8\% \\ 
\hline
Improvement&-&\textcolor{blue}{+2.4\%}&\textcolor{blue}{+3.2\%}&\textcolor{blue}{+3.5\%}&\textcolor{blue}{+4.8\%}&\textcolor{blue}{+3.4\%}\\
\hline
\hline

\multirow{2}{*}{MOTA} 
& Y & \textbf{91.4\%} & \textbf{90.3\%} & \textbf{86\%} & \textbf{86.8\%} & \textbf{85.3\%} \\ 
& N & 91\% & 89.4\% & 84.8\% & 85.5\% & 83.9\% \\ 
\hline
Improvement&-&\textcolor{blue}{+0.4\%}&\textcolor{blue}{+0.9\%}&\textcolor{blue}{+1.2\%}&\textcolor{blue}{+1.3\%}&\textcolor{blue}{+1.4\%}\\
\hline
\hline

\multirow{2}{*}{IDSW} 
& Y & \textbf{1} & \textbf{0} & \textbf{3} & \textbf{1} & \textbf{5} \\ 
& N & 2 & 5 & 10 & 14 & 10 \\ 
\hline
Improvement&-&\textcolor{blue}{-1}&\textcolor{blue}{-5}&\textcolor{blue}{-7}&\textcolor{blue}{-13}&\textcolor{blue}{-5}\\
\hline
\hline
\multirow{2}{*}{MT} 
& Y & \textbf{159} & \textbf{162} & \textbf{152} & \textbf{153} & \textbf{155} \\ 
& N & 159 & 160 & 146 & 151 & 151 \\ 
\hline
Improvement&-&\textcolor{blue}{0}&\textcolor{blue}{+2}&\textcolor{blue}{+6}&\textcolor{blue}{+2}&\textcolor{blue}{+4}\\
\hline
\end{tabular}
\end{adjustbox}
\end{table}
In this experiment, the performance of the KF without the proposed refinement for noise mitigation, $D_t$, is compared with the proposed refined KF. The experiment is conducted on five different detectors to justify the contribution of the proposed refinement KF, as demonstrated in Table~\ref{tab:ref_analysis}. The comparison focuses on metrics MOTA, HOTA, IDSW, and the number of mostly tracked objects (MT). For this ablation study, the noise mitigation term $D_t$ is modeled on the KITTI training dataset to measure the generalization of the mitigation term on the unseen dataset, which is the KITTI evaluation dataset in this case. Since this study aims to evaluate the impact of localization and noise mitigation on state estimation, Table~\ref{tab:ref_analysis} compares state prediction for objects \textbf{(Before refining the state given the new measurement)} and the actual location of the objects from the ground truth. Thus, the improvement row in Table~\ref{tab:ref_analysis} reflects enhancement in state prediction due to including the noise mitigation term $D_t$. 
\\
The proposed refinement in the KF consistently enhances the tracking performance across the five detectors, as shown in Table~\ref{tab:ref_analysis}. Notably, the improvement increases as the detection performance decreases. Table~\ref{tab:ref_analysis} shows significant improvement in HOTA reaches $4.8\%$ with PV-RCNN~\cite{point_rcnn_detector} detector. The primary reason for the significant improvement in HOTA compared to MOTA is that HOTA includes localization accuracy, while MOTA considers identity switch between objects (See Section~\ref{sec:implementation}). This observation shows the contribution of the proposed noise mitigation term $D_t$ that leads to enhanced state prediction for objects. The improvement in state prediction allows recovering objects that reach to recover about 13 objects with PV-RCNN~\cite{point_rcnn_detector} detector (See IDSW in Table~\ref{tab:ref_analysis}).
\\ These results, including the qualitative results represented in this work, confirm the contribution of the proposed KF refinement that mitigates the noise associated with the detections, enhancing state estimation and tracking performance.\\

\subsubsection{Track Validity and Ghost Tracks}
\label{subusbsec:track_val_and_gho_trak}
\begin{figure}[!tbh]
    \centering
    \includegraphics[width=7cm, height=14cm]{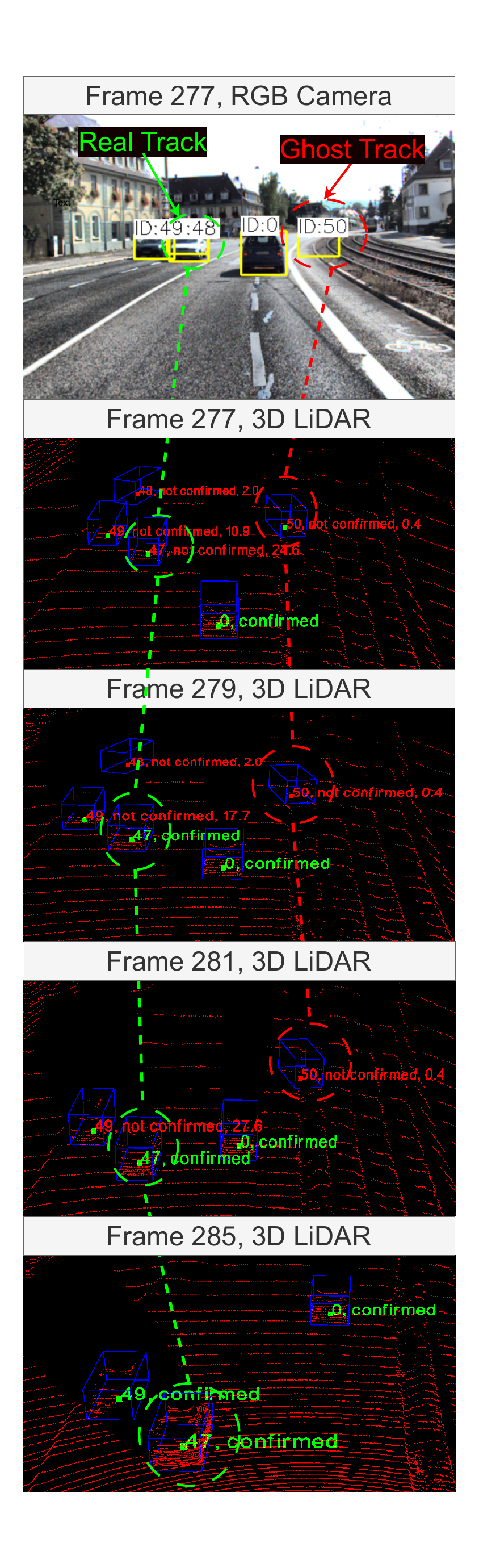}
    \caption{This scenario illustrates the proposed trajectory validity mechanism's capability of distinguishing between real tracks (Car ID:47, Car ID:49) and ghost tracks (Car ID:48, Car ID: 50). It showcases the ability to accurately identify and maintain a 'Not Confirmed' status for the ghost track while recognizing Car ID:49 and Car ID:47 as legitimate tracks.}
    \label{fig:ghost_real_comp}
\vspace{-0.4cm}
\end{figure}

The second concern raised in this research is ghost tracks. Ghost tracks are a form of consecutive false positive observations that generate illusion trajectories that could mislead MOT methods. This work proposes the first online trajectory validation mechanism to identify ghost tracks among legitimate trajectories in MOT methods. The contribution of the proposed mechanism is validated in two steps. First, a qualitative scenario is demonstrated in which ghost and legitimate tracks exist and the capability of the mechanism to identify legitimate tracks. Second, a sensitivity analysis is performed by disabling the mechanism and quantifying the improvement in the tracking performance. \\
Figure~\ref{fig:ghost_real_comp} reports a 
qualitative experiment of the trajectory validation mechanism for identifying legitimate tracks. The figure shows a scenario of three legitimate tracks for three cars with ID $49$, $48$, and $0$ and two ghost tracks with ID $50$ and $48$ at frame $277$. A car with ID $47$ is identified as a legitimate track at frame $279$, while the car with ID $49$ is confirmed at frame $285$. The ghost tracks with ID $49$, $48$, and $0$ remain unconfirmed until discarded from the system. This scenario shows the capability of the mechanism to identify legitimate tracks among ghost tracks.

\begin{table}[!htb]
\caption{\footnotesize Performance evaluation of our framework with and without the proposed trajectory validity mechanism across five detectors}
\label{tab:w_wo_track_val}
\centering
\begin{adjustbox}{max width=\columnwidth}
\begin{tabular}{@{}|l|c|c|c|c|@{}}
\toprule
 \textbf{Detector} & \textbf{Traj. Validity} &\textbf{HOTA $\uparrow$} & \textbf{MOTA $\uparrow$} & \textbf{IDSW $\downarrow$} \\ 
\midrule
PointRCNN~\cite{point_rcnn_detector} &N& 69.37\% &  58.54\% & 16\\ 
PointRCNN~\cite{point_rcnn_detector} &Y& \textbf{78\%} & \textbf{86.55\%} & \textbf{3}\\ 
\hline
Improvement &-& \textcolor{blue}{+8.63\%} & \textcolor{blue}{+28.01\%}&\textcolor{blue}{-13} \\
\midrule
PV-RCNN~\cite{pv-rcnn_detector} &N& 74.58\% &71.02\% & 35 \\
PV-RCNN~\cite{pv-rcnn_detector} &Y& \textbf{79.92\%} & \textbf{86.93\%} & \textbf{1}  \\ 
\midrule
Improvement &-&\textcolor{blue}{+5.34\%} &\textcolor{blue}{+15.91\%}& \textcolor{blue}{-34}\\
\midrule
Second~\cite{second_detector} &N& 69.80\% & 56.26\% & 30 \\ 
Second~\cite{second_detector} &Y& \textbf{78.73\%} & \textbf{85.73\%} & \textbf{7}\\ 
\midrule
Improvement  &-& \textcolor{blue}{+8.93\%} & \textcolor{blue}{+29.47\%}&\textcolor{blue}{-23} \\
\midrule
CasA~\cite{casA_detector} &N& 81.72\% & 86.13\% & 10 \\ 
CasA~\cite{casA_detector} &Y& \textbf{83.61\%} & \textbf{90.58\%} & \textbf{0} \\ 
\midrule
Improvement &-&\textcolor{blue}{+1.89\%} &\textcolor{blue}{+4.45\%}& \textcolor{blue}{-10}\\
\midrule
VirConv~\cite{VirConvDet_CVPR_23} &N& 85.87\% & 90.88\% & 10 \\ 
VirConv~\cite{VirConvDet_CVPR_23} &Y& \textbf{86.31\%} & \textbf{91.53\%} & \textbf{1}  \\
\hline
Improvement &-&\textcolor{blue}{+0.44\%} &\textcolor{blue}{+0.65\%}& \textcolor{blue}{-9}\\

\bottomrule
\end{tabular}
\end{adjustbox}

\end{table}

To support the proposed trajectory validity mechanism's impact on tracking performance, the mechanism is turned off and quantitatively evaluates the tracking performance on the KITTI evaluation dataset after fine-tuning the mechanism on the training dataset. Five different detectors are employed to demonstrate the generalization of the proposed mechanism and the performance enhancement across five detectors is quantified when the trajectory validity mechanism is enabled. In this study, the trajectory validity mechanism threshold $\sigma_{legit}$ is initially finetuned on the KITTI training dataset, and its performance is evaluated on the KITTI evaluation dataset. The evaluation is conducted by turning off the proposed mechanism and employing threshold on the detection score as proposed and finetuned in CasTrack~\cite{CasTrack} work. Next, the impact of the proposed mechanism can be observed when enabled.\\
The evaluation involves HOTA, MOTA, and IDSW metrics to represent the overall tracking performance. Table~\ref{tab:w_wo_track_val} shows that enabling the trajectory validity mechanism improves HOTA across all detectors, with a margin reaching $8.94\%$ with Second~\cite{second_detector} detector. Moreover, there is a significant improvement in the MOTA metric for overall detectors that peak with the PointRCNN detector with up to $28.01\%$ enhancement in MOTA. Since the trajectory validity mechanism identifies legitimate tracks while allowing their observations to pass through the system, the tracking performance becomes more consistent with less misidentification for objects that reflect few IDSW values, as shown in Table~\ref{tab:w_wo_track_val}. It results in a reduction in IDSW reaches $34$ with PV-RCNN~\cite{pv-rcnn_detector} detector.\\

The combination of the qualitative results in Figure~\ref{fig:ghost_real_comp} and the quantitative results in Table~\ref{tab:w_wo_track_val} confirms the contribution of the proposed mechanism to object tracking performance.

\section{Limitations and Future Work}
This work observes trajectory drift for occluded objects, which fails MOT approaches to recover these objects after the end of the occlusion. This research shows that this drift results from noise from the alimitation and futuressigned deep-learning detector that impacts the localization of the detected objects. Accumulating this noise shapes a Gaussian distribution that facilitates the integration with the KF. The proposed term mitigates the need for drift-enhancing state estimation for objects, which leads to recovering objects under extended occlusion scenarios. Nevertheless, this term assumes the impact of this noise on objects is almost equal, even though this noise can differ as per the view perspective of the object, including the number of observations for that object. Although the error that comes from that assumption is partially filtered during state estimation through the KF, dismissing that assumption by a dynamic formulation for this noise can allow further enhancement in state estimation. \\
Another contribution introduced in this work is to overcome the formulation of ghost tracks in MOT by replacing the traditional threshold-based filter approach with a trajectory validity mechanism that provides real-time distinction between legitimate and ghost tracks in MOT approaches. As shown in the ablation study in Section~\ref{subusbsec:track_val_and_gho_trak}, the mechanism significantly enhances tracking performance in the MOTA metric, which indicates a reduction in false-positive tracks in the method. Moreover, this reduction in false tracks prevents further mismatching for objects' identities showing less IDSW, as presented in Table~\ref{tab:kitt_test_eval}.  Nevertheless, the mechanism requires multiple observations to validate the legitimacy of a track, which means this constraint may exclude legitimate tracks for objects with too fewer observations, i.e., only one or two observations for the object. Our observation of this limitation concludes that most of these objects are already leaving the scene of the AV; keeping their track in the MOT method does not provide important information for the AV. Nevertheless, this observation triggers the question, "Which objects should be tracked?". Current research in MOT assumes that all objects must be tracked; however,  this is not always true for real-time AV environments. For instance, tracking information for cars driving on a road different than the AV is not always beneficial for the AV to navigate safely, but overloading the MOT method instead can eventually harm the tracking performance of other necessary tracks.\\
Lastly, this work uses Kalman with a constant acceleration motion model for state estimation; however, this assumption does not apply to all objects. For instance, vehicles regularly change their motion according to traffic. This assumption also impacts the precision of state estimation for objects in the KF.

\section{Conclusion}
This paper addresses two key challenges in multi-object tracking (MOT): state estimation deviation and ghost tracks. This research finds that spatial localization noise from deep learning detectors impacts bounding box localization accuracy, leading to imprecise motion parameter estimation and trajectory drift. This issue is observed in multiple scenarios, including the recent benchmarks. Moreover, this research discusses the limitation of the current approach followed in the literature, threshold-based filtration, to filter out ghost tracks (false positives) and its limitation of leaking ghost tracks to the MOT approach while blocking legitimate ones. This work tackles these limitations by introducing a refined Kalman filter that compensates for the spatial localization noise, enabling trajectory drift mitigation and recovery of prolonged occluded objects. In addition, The work introduces a novel online trajectory validity mechanism to identify legitimate tracks based on ghost track characteristics. Finally, the work proposes an online MOT framework, RobMOT, which outperforms state-of-the-art trackers on the KITTI and Waymo Open Dataset (WOD). RobMOT shows superior tracking performance across various detectors, with  HOTA improvements up to 3.92\% and 8.7\% on the KITTI testing and validation dataset. Compared to SOTA methods, the framework excels in tracking distant and recovering occluded objects with a distance exceeding 50 meters, followed by a 1.27\% improvement in MOTA on the WOD testing split. The framework shows computational efficiency that reflects its applicability for real-time applications.  
\appendix
\begin{table}[htb]
\caption{\footnotesize List of Thresholds and Parameter Values}
\label{tab:thresholds}
\centering
\begin{adjustbox}{max width=\columnwidth}
\begin{tabular}{@{}|l|c|c|c|c|c|c|@{}}
\toprule
 \textbf{Detector} & $D_t(\sigma_x^2)$ &  $D_t(\sigma_y^2)$ & $\sigma_{nconf}$&$\sigma_{conf}$&$\sigma_{legit}$&$\sigma$\\ 
 \midrule
 VirConv~\cite{VirConvDet_CVPR_23} &0.017221&0.005901&0&-1&20&4\\ 
  \midrule
 CasA~\cite{casA_detector} &0.034966& 0.019720& 0 & 0& 25&3\\ 
   \midrule
PointRCNN~\cite{point_rcnn_detector} &0.030874&0.009379&0&0&35&4\\
\midrule
PV-RCNN~\cite{pv-rcnn_detector}&0.036383&0.013067&0.5&0.5&20&2\\
\midrule
Second~\cite{second_detector}&0.039156&0.014357&-1&-2&10&3\\
%


\bottomrule
\end{tabular}
\end{adjustbox}
\end{table}
The trajectory is discarded from the cache for all detectors when the estimation uncertainty $diag(P_t)_x$ or $diag(P_t)_y$ exceeds $\sigma_{est\_certainity} = 4$ meters. 
 \bibliographystyle{unsrt} 
\bibliography{main}
\end{document}